\def\paperTitle{
DiD It in 87 Minutes:\\
A Label-Free Softmax-to-Linear Adaptation\\
of Vision Transformers for Object Detection}

\def\authorBlock{
{\bfseries\fontsize{11.2}{16}\selectfont
    Huaiyuan Qin\,\textnormal{\textsuperscript{1,*}}%
\ \ \ \hspace{24pt}
    Gabriel James Goenawan\,\textnormal{\textsuperscript{1,*,$\dagger$}}%
\ \ \ \hspace{24pt}
    Zihang Lin\,\textnormal{\textsuperscript{1,2,*,$\dagger$}}%
}\vspace{3pt}\\

{\bfseries\fontsize{11.2}{16}\selectfont
    Muli Yang\,\textnormal{\textsuperscript{1}}%
\ \ \ \hspace{24pt}
    Hongyuan Zhu\,\textnormal{\textsuperscript{1,\Letter}}%
}
\vspace{4pt}
\\ 

{\fontsize{10.9}{16}\selectfont
\textsuperscript{1}Institute of Advanced Intelligence and Computing (IAIC), A*STAR, Singapore
}\\

{\fontsize{10.9}{16}\selectfont\textsuperscript{2}Nanyang Technological University}
\vspace{3pt}\\

{\tt
\small
\{qinhy, goenawan, yangml, zhuh\}@a-star.edu.sg,
linz0071@e.ntu.edu.sg
\hspace{-6.28pt}}\\
}

\newif\ifreview 
\newif\ifarxiv \newcommand{\arxiv}{\arxivtrue}
\newif\ifcamera 
\arxiv

\pdfoutput=1
\documentclass{article}

\PassOptionsToPackage{numbers, compress}{natbib}
 
\ifreview \usepackage{meta/neurips_2026} \fi
\ifarxiv \usepackage[preprint]{meta/neurips_2026} \fi
\ifcamera \usepackage[main, final]{meta/neurips_2026} \fi

\usepackage{fix-cm}
\usepackage{array}
\usepackage{nicematrix}

\usepackage{tipa}
\usepackage{dsfont}
\usepackage{etoolbox}  

\usepackage[noend]{algorithmic}
\usepackage{algorithm}

\usepackage{float}
\usepackage{newfloat}
\usepackage{listings}

\usepackage{subcaption}
\floatstyle{ruled}
\newfloat{listing}{tb}{lst}{}
\floatname{listing}{\small Algorithm}

\definecolor{mygray}{RGB}{234,234,234}

\usepackage{adjustbox}
\usepackage[table,dvipsnames]{xcolor}

\definecolor{darkgreen}{rgb}{0.13, 0.55, 0.13}

\usepackage{graphicx}	
\usepackage{amsmath}	
\usepackage{amssymb}	
\usepackage{booktabs}
\usepackage{times}
\usepackage{epsfig}
\usepackage{caption}
\usepackage{float}
\usepackage{placeins}
\usepackage{color, colortbl}
\usepackage{enumitem}
\usepackage{tabularx}
\usepackage{xstring}
\usepackage{multirow}
\usepackage{xspace}
\usepackage{subcaption}
\usepackage{xcolor}

\usepackage{inconsolata}

\ifcamera \usepackage[accsupp]{axessibility} \fi

% ...

\ifarxiv  \fi

\newcommand{\R}[1]{{%
    \textbf{%
        \ifstrequal{#1}{1}{\textcolor{red}{R#1}}{%
        \ifstrequal{#1}{2}{\textcolor{blue}{R#1}}{%
        \ifstrequal{#1}{3}{\textcolor{magenta}{R#1}}{%
        \ifstrequal{#1}{4}{\textcolor{teal}{R#1}}{%
                           \textcolor{cyan}{R#1}%
        }}}}%
    }%
}}

\definecolor{Gray}{gray}{0.5}
\definecolor{nicergreen}{rgb}{0.13, 0.54, 0.13}
\definecolor{nicered}{rgb}{0.83, 0.16, 0.16}
\definecolor{lightgray}{RGB}{230, 230, 230}
\definecolor{Highlight}{HTML}{39b54a}  %

\usepackage{appendix}
\usepackage{footnote}

\usepackage{microtype}
\usepackage{cuted}
\usepackage{tocloft}

\usepackage[T1]{fontenc}
\usepackage{DejaVuSans}

\usepackage{ifsym, marvosym}

\let\svthefootnote\thefootnote
\newcommand\freefootnote[1]{%
  \let\thefootnote\relax%
  \footnotetext{#1}%
  \let\thefootnote\svthefootnote%
}

\makeatletter
\DeclareRobustCommand\onedot{\futurelet\@let@token\@onedot}
\def\@onedot{\ifx\@let@token.\else.\null\fi\xspace}

\makeatother

\usepackage{wrapfig}

\usepackage{xr-hyper}

\makeatletter
\newcommand*{\addFileDependency}[1]{
  \typeout{(#1)}
  \@addtofilelist{#1}
  \IfFileExists{#1}{}{\typeout{No file #1.}}
}

\makeatother

\definecolor{cvprblue}{rgb}{0.21,0.49,0.74}
\usepackage[pagebackref,breaklinks,colorlinks]{hyperref}
\usepackage[capitalize]{cleveref}
\crefname{section}{Sec.}{Secs.}
\crefname{table}{Table}{Tables}
\crefname{figure}{Fig.}{Figs.}

\usepackage[utf8]{inputenc} 
\usepackage[T1]{fontenc}    
\usepackage{url}            
\usepackage{booktabs}       
\usepackage{amsfonts}       
\usepackage{amssymb}        
\usepackage{nicefrac}       
\usepackage{microtype}      
\usepackage{xcolor}         

\begin{document}
\title{\paperTitle}
\author{\authorBlock}
\maketitle

\ifarxiv
\freefootnote{\hspace{-12pt}\textsuperscript{*} Equal first author. Authors are permitted to list their name first in their CVs.}
\freefootnote{\hspace{-12pt}\textsuperscript{$\dagger$} This paper was completely accomplished when GJG worked and ZL interned at A*STAR.}
\freefootnote{\hspace{-12pt}\textsuperscript{\Letter} Corresponding author.}
\fi

\begin{abstract}
While linear attention is a compelling mechanism for high-resolution object detection due to its reduced cost for global token mixing, converting the Softmax-attention ViT backbone of a trained detector into a linear-attention one is not a trivial drop-in replacement. 
Directly swapping the attention operator leads to severe performance degradation, and generic label-free distillation, though effective for classification, often fails on detection tasks. 
We argue that the central challenge is \emph{detector-interface preservation}: the converted backbone must reproduce the exact feature tensors expected by the fixed downstream detector, rather than merely imitating internal Softmax hidden states. 
To address this, we introduce Detector-Interface Distillation (DiD), a label-free conversion method that exclusively trains the linear-attention backbone by aligning detector-facing interface tensors with those of a frozen Softmax teacher. 
On DOTA-v1.5, DiD substantially outperforms established baselines and matches supervised, fully trained linear models. Adaptation completes in roughly 87 minutes on 4 GPUs, and the linearized backbone cuts inference latency by $\sim$62\% and peak memory by $\sim$49\%. 
We hope our findings offer the community a simple, label-free route to reusing trained Softmax detectors as efficient linear ones, and encourage interface-aware objectives in future architecture-conversion work.

\end{abstract}

\section{Introduction}

Vision Transformers (ViTs) have become a dominant backbone family in modern computer vision.
Their strong performance is driven by scalable self-attention~\cite{zhai2022scaling,dehghani2023scaling,fang2023eva}, large-scale pretraining~\cite{singh2023effectiveness,yang2025scaling,wang2025scaling}, and effective transfer to downstream tasks such as image classification~\cite{dosovitskiy2020image}, semantic segmentation~\cite{zhang2022segvit,kirillov2023segment,ravi2025sam,carion2025sam}, and object detection~\cite{fang2021you,zhang2021vit,li2022exploring,chen2022vision}.
In detection, ViT-based systems such as ViTDet~\cite{li2022exploring} and transformer detectors such as YOLOS~\cite{fang2021you} show that pure or nearly pure transformer backbones can provide strong representations when paired with an appropriate detector head or feature pyramid.
However, this success comes with a fundamental limitation: standard Softmax attention scales quadratically with the number of image tokens~\cite{keles2023computational}.
This cost becomes especially problematic for dense prediction, where high-resolution inputs are often required to localize small objects~\cite{wang2021pyramid,liu2021swin,yang2021focal,yuan2021hrformer,ranftl2021vision}, as in application scenarios like remote sensing~\cite{xia2018dota} and autonomous driving~\cite{yu2020bdd100k}.

To address this cost, a large body of work has explored more efficient ViT designs.
One line of research reduces attention complexity through locality or hierarchy~\cite{wang2021pyramid,liu2021swin} that restricts self-attention to local windows or progressively aggregated regions, substantially improving efficiency while maintaining strong dense-prediction performance.
These hierarchical backbones have become highly practical for detection and have been adopted in strong detector families~\cite{zhang2022dino,zong2023detrs,liu2024grounding,ren2024dino}.
Nevertheless, hierarchy and locality also impose architectural constraints: long-range interactions become indirect~\cite{yang2021focal,liu2022swin,ding2022davit}, and the model no longer preserves the same global token-mixing behavior as a standard Softmax-attention ViT.

\begin{figure*}[t]
  \centering
  \includegraphics[width=0.76\textwidth]{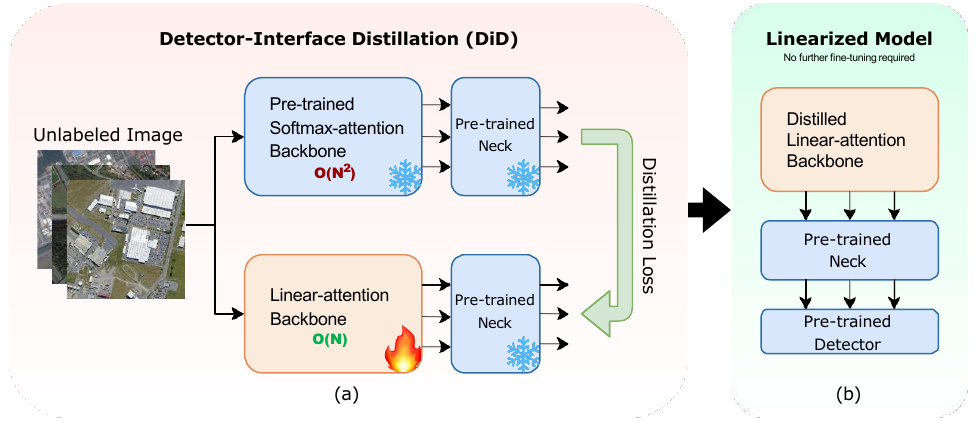}
  \vspace{-6pt}
  \caption{\textbf{Overview of the label-free Softmax-to-linear detector conversion setting.}
  (a) Starting from a trained detector with a Softmax-attention ViT backbone, we construct a student by replacing the backbone attention operator with linear attention.
  During conversion, the Softmax teacher and the linear student receive the same unlabeled images; only the student backbone is updated, while the downstream detector remains fixed.
  Snowflake symbols indicate frozen modules, while the fire symbol indicates the trainable module.
  (b) After conversion, the distilled linear backbone is used directly in the detector without additional fine-tuning.}
  \vspace{-12pt}
  \label{fig:did_overview}
\end{figure*}

Another line of work targets the quadratic attention operator itself~\cite{katharopoulos2020transformers,han2024bridging,zhang2024hedgehog,zheng2025linear}, inspired by earlier work in NLP~\cite{2020arXiv200104451K,choromanski2020rethinking,qin2022cosformer}.
Linear-attention methods replace the Softmax kernel with feature-map-based formulations whose complexity scales linearly with sequence length.
Early linear-attention variants often trailed quadratic Softmax attention in accuracy~\cite{katharopoulos2020transformers}, but more recent approaches~\cite{han2024bridging,zhang2024hedgehog,zheng2025linear} have narrowed this gap and, in some settings, surpassed quadratic attention.
These results suggest that linear attention can be competitive under suitable architectures and training regimes.
However, adopting linear attention in an existing detector typically demands pre-training the linear backbone from scratch and then fine-tuning the entire detector to recover accuracy, both of which require substantial compute in addition to labeled data.

This creates a practical conversion problem: \textbf{can a quadratic-attention backbone in an existing object detector be converted into a linear-attention backbone without labels and without retraining the entire detector?}

In this paper, we study this problem in the setting of lightweight object-detection conversion.
Rather than retraining the full detector, we replace the quadratic attention modules in the backbone of a trained ViT detector with linear attention and perform a short label-free adaptation stage that trains only the backbone, as illustrated in Figure~\ref{fig:did_overview}.
The difficulty is that the converted backbone must remain compatible with a detector pipeline that was trained around the original Softmax feature distribution.

We show that this is not merely a generic representation-transfer problem.
For classification, preserving global semantic behavior is often sufficient~\cite{tung2019similarity,tian2019contrastive}, but object detection is more restrictive because downstream performance depends on the deployed multi-scale interface tensors consumed by the detector~\cite{guo2021distilling,golizadeh2025architectural}.
Prior detector distillation works suggest that supervision at intermediate detector features can be particularly effective for object detection~\cite{guo2021distilling,golizadeh2025architectural}.
We build on this general intuition by preserving the detector-facing feature contract required by the fixed downstream detector after the attention operator is changed.
Based on this observation, we propose \emph{Detector-Interface Distillation} (DiD), a label-free conversion method that supervises detector-facing interface tensors rather than the full sequence of hidden states.
Across our experiments, this simple detector-aware objective substantially improves Softmax-to-linear transfer for detection and yields a practical adaptation procedure that completes in roughly 87 minutes on 4 GPUs.

Our contributions are as follows:

\begin{enumerate}
\item We identify detector-interface mismatch as a key failure mode in Softmax-to-linear attention conversion for object detection.
Direct attention replacement collapses performance, and generic label-free distillation remains insufficient though it preserves classification behavior well.
\item We propose Detector-Interface Distillation (DiD), a simple label-free conversion method that aligns detector-facing interface tensors while keeping the downstream detector fixed.
\item We provide a controlled study showing that detector-interface supervision substantially outperforms direct copy, generic self-supervised distillation, and hidden-state distillation, and is competitive with full supervised training of the linear detector.
\item We quantify the practical payoff of conversion, showing that the resulting ReLU backbone substantially reduces latency and memory while the conservative $40$k-step adaptation completes in about $87$ minutes on 4 GPUs.
\end{enumerate}

%
\section{Related Work}

\paragraph{Efficient Attention for Vision.}
The quadratic complexity of Softmax attention has motivated several families of efficient visual architectures.
Local and hierarchical transformers restrict attention to windows or progressively aggregated regions~\cite{wang2021pyramid,liu2021swin,yang2021focal,ding2022davit}, while sparse designs reduce the number of token interactions.
These approaches are highly effective for dense prediction, but long-range information is propagated indirectly and the resulting architecture differs substantially from a standard global-attention ViT.
Linear-attention methods instead modify the attention operator itself~\cite{katharopoulos2020transformers,qin2022cosformer,han2024bridging,zhang2024hedgehog,zheng2025linear}: replacing Softmax with a feature-map formulation and reordering the matrix multiplication reduces the attention complexity from quadratic to linear in sequence length.
These designs have narrowed the accuracy gap to Softmax attention, making linear attention a promising option for high-resolution vision.
More recent attempts reuse pre-trained Softmax weights to build linear-attention ViTs~\cite{li2026vit,li2026linearizing}, but still rely on multi-stage adaptation or full retraining.
Our work is complementary: rather than designing or training a new linear-attention architecture from scratch, we study how to convert the backbone of an already trained detector without retraining the detection pipeline.

\paragraph{Label-Free Model Conversion.}
Knowledge distillation transfers behavior from a frozen teacher to a student using teacher-derived supervision~\cite{hinton2015distilling,gou2021knowledge}.
In label-free settings, common objectives match logits, global representations, hidden states, attention outputs, or attention statistics.
Softmax-to-linear conversion methods similarly treat operator replacement as a distillation problem, matching hidden states~\cite{goldstein2025radlads}, attention behavior~\cite{zhang2024hedgehog}, or low-rank approximations of attention maps~\cite{zhang2025lolcats}.
Related operator-conversion pipelines have also been explored for language models~\cite{kasai2021finetuning,wang2024mamba}.
Most existing conversion methods assume that any remaining mismatch can be repaired through subsequent supervised fine-tuning.
Our setting removes this recovery stage: the converted backbone must be immediately compatible with a frozen detector, without detection labels.
This changes the objective from approximate operator imitation to preservation of the deployed task interface.

\paragraph{Knowledge Distillation for Object Detection.}
Detection distillation has shown that intermediate feature supervision can be more effective than output-level imitation because detectors depend on structured spatial features and strongly imbalanced foreground-background signals~\cite{guo2021distilling,yang2022focal,golizadeh2025architectural}.
Conventional detector distillation, however, generally transfers a larger teacher into a smaller student, often with detection labels and while training the detector pathway.
DiD addresses a different setting: teacher and student have comparable capacity, the attention operator is changed post hoc, the downstream detector remains fixed, and conversion uses only unlabeled images.
We build on the general importance of detector features but target the specific interface compatibility required by architectural conversion.
Our setting is also related to attention-transfer studies~\cite{li2024attention,qin2026attention}, which transfer the attention behavior of a pre-trained ViT to a student model; DiD instead preserves the detector-facing interface after the attention operator itself is replaced.

\section{Problem Formulation and Diagnostic Study}
\label{sec:motivation}

\begin{table*}[ht]
\centering
\small
\setlength{\tabcolsep}{3pt}
\begin{tabular}{@{}lcccccc@{}}
\toprule
& \multicolumn{4}{c}{\textbf{Classification}}
& \multicolumn{2}{c}{\textbf{Detection}} \\
\cmidrule(lr){2-5}\cmidrule(lr){6-7}
\multirow{2}{*}{Method}
& \multicolumn{2}{c}{ImageNet}
& \multicolumn{2}{c}{CIFAR-100}
& \multicolumn{2}{c}{DOTA-v1.5} \\
\cmidrule(lr){2-3}\cmidrule(lr){4-5}\cmidrule(lr){6-7}
& Acc. & Retained (\%)
& Acc. & Retained (\%)
& mAP & Retained (\%) \\
\midrule
Softmax reference
& 79.83 & 100.0
& 78.09 & 100.0
& 33.5  & 100.0 \\
\midrule
Direct copy
& 5.08  & 6.4
& 3.54  & 4.5
& 1.7   & 5.1 \\
Generic distillation
& 77.83 & 97.5
& 75.92 & 97.2
& 18.1  & 54.0 \\
\bottomrule
\end{tabular}
\caption{
\textbf{Cross-task Softmax-to-ReLU transfer.}
Each task reports its native score followed by the percentage retained relative to its own Softmax reference.
Generic distillation nearly restores classification accuracy but recovers only half of the DOTA-v1.5 detector mAP.
}
\vspace{-6pt}
\label{tab:cross_task}
\end{table*}

\subsection{Label-Free Softmax-to-Linear Conversion}
\label{sec:conversion_setting}

For query, key, and value projections $Q$, $K$, and $V$, standard Softmax attention is
\begin{equation}
\mathrm{Attention}_{\mathrm{Softmax}}(Q,K,V)
=
\mathrm{Softmax}\left(\frac{QK^\top}{\sqrt{d}}\right)V,
\end{equation}
which explicitly forms a token-to-token affinity matrix.
Linear attention replaces this operation with an associative feature-map form,
\begin{equation}
\mathrm{Attention}_{\mathrm{linear}}(Q,K,V)
=
\phi(Q)\left(\phi(K)^\top V\right),
\end{equation}
where $\phi(\cdot)$ is a non-negative feature map.
This reordering avoids explicitly materializing the full affinity matrix and improves scaling with sequence length.
However, it also changes the backbone function: parameters trained jointly with Softmax attention are not guaranteed to preserve the teacher behavior after the operator is replaced.

Let $B_t$ denote the trained Softmax backbone and $B_s$ the linear-attention student initialized from all compatible teacher parameters.
We decompose the trained downstream detector pathway as
\begin{equation}
G = H \circ N,
\end{equation}
where $N$ denotes the neck or feature-pyramid transformation and $H$ denotes the subsequent prediction modules.
During conversion, $B_t$, $N$, and $H$ remain frozen, and only $B_s$ is optimized using unlabeled images $x\sim\mathcal{D}$:
\begin{equation}
\theta_s^\star
=
\arg\min_{\theta_s}
\mathbb{E}_{x\sim\mathcal{D}}
\left[
\mathcal{L}_{\mathrm{conv}}
\bigl(B_s(x),B_t(x);N,H\bigr)
\right].
\label{eq:conversion_problem}
\end{equation}
The central question is therefore not merely how closely $B_s$ imitates $B_t$, but which teacher-student discrepancy must be minimized to keep the downstream detector functional.

\subsection{Why Is Generic Transfer Insufficient?}
\label{sec:generic_transfer_gap}

To test whether a classification-oriented transfer objective is sufficient for object detection, we compare direct copy and generic LightlyTrain-style global-feature distillation on ImageNet and CIFAR-100 classification, and on DOTA-v1.5 detection with RetinaNet.
Because classification accuracy and detection mAP are not directly comparable, each task reports both its native metric and the percentage retained relative to its own Softmax reference.

Direct copy collapses across all three tasks (Table~\ref{tab:cross_task}), confirming that the attention operator cannot be replaced without adaptation.
Generic distillation, however, exhibits a clear task gap: it retains 97.5\% and 97.2\% of the Softmax-reference accuracy on ImageNet and CIFAR-100, respectively, but only 54.0\% of the reference mAP on DOTA-v1.5.
Preserving global semantic representations is therefore insufficient to maintain the structured, multi-scale features required by a fixed detector, motivating a supervision target tied to the deployed detection pathway.

\subsection{Where Should Conversion Be Supervised?}
\label{sec:motivation_placement}

The cross-task comparison suggests that preserving global semantic representations is insufficient for object detection.
We therefore ask where supervision should be applied so that the converted backbone remains compatible with the fixed detector.
Using RetinaNet on DOTA-v1.5, we conduct a controlled 20k-step conversion in which the reconstruction loss and optimization settings are fixed while only the supervised tensors are changed.
We compare five candidate targets: attention outputs, hidden states from all transformer layers, hidden states from the backbone layers read by the neck, the final hidden state, and the multi-scale features produced after the frozen neck/FPN and passed to the detector heads.
For a selected tensor set $\mathcal{V}$, we use the following loss:
\begin{equation}
\mathcal{L}_{\mathrm{place}}
=
\frac{1}{|\mathcal{V}|}
\sum_{v\in\mathcal{V}}
\left\lVert
v_s-v_t
\right\rVert_F^2,
\end{equation}
where $v_t$ and $v_s$ denote corresponding teacher and student tensors.

\begin{wraptable}[12]{r}{0.56\textwidth}
\centering
\small
\setlength{\tabcolsep}{3pt}
\vspace{-10pt}
\begin{tabular}{@{}p{4.7cm}ccc@{}}
\toprule
Supervision target & mAP & mAP${_{50}}$ & mAP${_{75}}$ \\
\midrule
Attention outputs & 7.8 & 17.4 & 5.6 \\
Hidden states from all layers & 25.2 & 44.7 & 24.6 \\
Hidden states from readout layers & 20.3 & 37.4 & 19.0 \\
Final hidden state & 12.6 & 24.1 & 11.7 \\
\textbf{Post-neck/FPN detector features}
& \textbf{30.7} & \textbf{54.1} & \textbf{30.1} \\
\bottomrule
\end{tabular}
\caption{
\textbf{Effect of supervision placement for RetinaNet conversion on DOTA-v1.5.}
Matching multi-scale features produced after the neck/FPN is substantially more effective than matching attention outputs or internal backbone states.
}
\vspace{-6pt}
\label{tab:ablation_location}
\end{wraptable}

As shown in Table~\ref{tab:ablation_location}, matching all transformer hidden states is the strongest detector-agnostic target, but it remains $5.5$ mAP below supervision after the neck/FPN.
Matching only the backbone layers supplied to the neck is also insufficient.
Therefore, proximity to the detector does not by itself guarantee compatibility; the converted backbone must reproduce the transformed multi-scale features that the detector actually receives.

This experiment identifies the feature boundary between the frozen neck/FPN and the detector heads as the appropriate conversion target.
We refer to this boundary as the \emph{detector-facing interface} and formalize its preservation as Detector-Interface Distillation in the next section.


\section{DiD: Detector-Interface Distillation}
\label{sec:did}

\subsection{Overview}

Detector-Interface Distillation (DiD) adapts a linear-attention student backbone to a detector that was originally trained with a Softmax-attention backbone.
The frozen teacher and the student process the same unlabeled image, and their backbone readout features are passed through identical frozen neck/FPN pathways.
DiD then aligns the multi-scale tensors produced at the output of the neck/FPN, while updating only the student backbone.

The key design is therefore not to reproduce the teacher's complete internal computation trace.
Softmax and linear attention implement different token-mixing rules, so their intermediate hidden states need not follow identical trajectories.
DiD instead preserves the feature boundary on which the fixed detector depends.
Internal transformer states and final detector predictions are not used as distillation targets.
Algorithm~\ref{algo:did} summarizes this computation in pseudocode.

\begin{algorithm}[tb]
   \caption{DiD PyTorch-style pseudocode.}
   \label{algo:did}
    \definecolor{codeblue}{rgb}{0.25,0.5,0.5}
    \lstset{
      basicstyle=\fontsize{7.2pt}{7.2pt}\ttfamily\bfseries,
      commentstyle=\fontsize{7.2pt}{7.2pt}\color{codeblue},
      keywordstyle=\fontsize{7.2pt}{7.2pt},
    }
\begin{lstlisting}[language=python]
# B_t: frozen Softmax teacher backbone
# B_s: trainable linear-attention student
# N_t, N_s: identical frozen neck/FPN copies
# H: frozen detector heads (unused in loss)
# beta: weights, beta[i] for interface m = i+1
B_s.params = B_t.params # copy compatible weights
for x in unlabeled_images: # no detection labels
    R_t = B_t.readout_features(x) # frozen path
    R_s = B_s.readout_features(x) # trainable

    Z_t = N_t(R_t) # tensors z_t^(1), ..., z_t^(M)
    Z_s = N_s(R_s) # tensors z_s^(1), ..., z_s^(M)

    # weighted detector-interface reconstruction
    loss = 0
    for i in range(M):
        loss += beta[i]*mse(Z_s[i], Z_t[i].detach())
    loss /= sum(beta)

    loss.backward() # update only student backbone
    update(B_s) # AdamW step

# deploy: drop teacher, reuse the frozen detector
converted_detector = H o N_s o B_s
\end{lstlisting}
\end{algorithm}

\subsection{Detector-Facing Interface}

Let $B_t$ denote the frozen Softmax teacher backbone and $B_s$ the trainable linear-attention student backbone.
For an input image $x$, the two backbones expose the readout sets
\begin{equation}
\mathcal{R}_t(x)
=
\left\{
r_t^{(\ell)}(x)
\right\}_{\ell\in\mathcal{I}},
\qquad
\mathcal{R}_s(x)
=
\left\{
r_s^{(\ell)}(x)
\right\}_{\ell\in\mathcal{I}},
\end{equation}
where $\mathcal{I}$ indexes the backbone layers read by the detector neck.
The exact readout locations depend on the detector architecture.

Let $N$ denote the pretrained neck or feature-pyramid pathway.
During conversion, the teacher and student use identical frozen copies of $N$.
Their detector-facing outputs are
\begin{equation}
\mathcal{Z}_t(x)
=
\left\{
z_t^{(m)}(x)
\right\}_{m=1}^{M}
=
N\left(\mathcal{R}_t(x)\right),
\end{equation}
and
\begin{equation}
\mathcal{Z}_s(x)
=
\left\{
z_s^{(m)}(x)
\right\}_{m=1}^{M}
=
N\left(\mathcal{R}_s(x)\right).
\end{equation}
We refer to $\mathcal{Z}_t(x)$ and $\mathcal{Z}_s(x)$ as the \emph{detector-facing interfaces}: they are the multi-scale feature tensors presented by the frozen neck/FPN to the downstream detector.
This definition separates the supervision target from the backbone readouts themselves.
The readout tensors $\mathcal{R}(x)$ are inputs to the neck/FPN, whereas the interface tensors $\mathcal{Z}(x)$ are its outputs and constitute the feature contract consumed by the detector.

DiD supervises this post-neck/FPN interface rather than every internal hidden state.
It also does not distill final detector predictions, because doing so introduces detector-specific prediction, matching, and assignment logic beyond the feature interface itself.
In the RetinaNet setting used for the supervision-placement and objective-design studies, $M=5$.

\vspace{-3pt}
\subsection{Weighted Interface Objective}

DiD aligns corresponding teacher and student interface tensors using the weighted reconstruction objective
\begin{equation}
\mathcal{L}_{\mathrm{DiD}}(x)
=
\frac{1}{\sum_{m=1}^{M}\beta_m}
\sum_{m=1}^{M}
\beta_m
\left\lVert
z_s^{(m)}(x)-z_t^{(m)}(x)
\right\rVert_F^2,
\label{eq:did_loss}
\end{equation}
where $\beta_m>0$ controls the contribution of interface output $m$.
For the five-interface RetinaNet setting, the default configuration uses $(\beta_1,\ldots,\beta_5)=(3.0,\,2.5,\,2.0,\,1.5,\,1.0)$, following the early-to-late output ordering used in the implementation.
We use plain MSE because it directly preserves the absolute feature scale and channel responses expected by the frozen detector.
The interface-weighting and loss-design comparisons are reported in Section~\ref{sec:ablations}.

\vspace{-3pt}
\subsection{Conversion and Deployment}

Recall that $N$ denotes the frozen neck/FPN and $H$ denotes the frozen detector prediction modules that consume its outputs.
Given an unlabeled conversion set $\mathcal{D}$, DiD optimizes only the student-backbone parameters:
\begin{equation}
\theta_s^\star
=
\arg\min_{\theta_s}
\mathbb{E}_{x\sim\mathcal{D}}
\left[
\mathcal{L}_{\mathrm{DiD}}(x)
\right].
\label{eq:did_optimization}
\end{equation}
For each image, the teacher provides the detached targets $\mathcal{Z}_t(x)$ and the student produces $\mathcal{Z}_s(x)$.
The teacher backbone, both neck/FPN pathways, and the detector prediction modules remain frozen.
DiD uses no ground-truth boxes, class labels, detector-output targets, or pseudo-label assignment.

After conversion, the teacher branch is discarded.
The converted detector becomes $f_s^\star = H\circ N\circ B_s^\star$, which reuses the original neck/FPN and prediction modules without additional fine-tuning.

\section{Experiments}
\label{sec:experiments}

\begin{table*}[t]
\centering
\small
\setlength{\tabcolsep}{3pt}
\begin{tabular}{llccc|ccc}
\toprule
\multirow{2}{*}{Detector} & \multirow{2}{*}{Method}
& \multicolumn{3}{c}{ReLU}
& \multicolumn{3}{c}{ELU} \\
& & mAP & mAP${_{50}}$ & mAP${_{75}}$ & mAP & mAP${_{50}}$ & mAP${_{75}}$ \\
\midrule
\multirow{7}{*}{RetinaNet}
& \cellcolor[HTML]{d8d8d8}Softmax reference & \cellcolor[HTML]{d8d8d8}33.5 & \cellcolor[HTML]{d8d8d8}58.1 & \cellcolor[HTML]{d8d8d8}33.2 & \cellcolor[HTML]{d8d8d8}33.5 & \cellcolor[HTML]{d8d8d8}58.1 & \cellcolor[HTML]{d8d8d8}33.2 \\
& Full training & 30.9 & 53.3 & 30.9 & 30.8 & 53.0 & \textbf{31.3} \\
\cmidrule{2-8}
& Direct copy & 1.7 & 3.8 & 1.2 & 1.3 & 2.9 & 0.9 \\
& Generic distillation & 18.1 & 35.2 & 16.6 & 9.6 & 19.2 & 5.3 \\
& Softmax-generated pseudo-label distillation
& 14.8 & 33.6 & 10.1 & 19.6 & 38.2 & 17.2 \\
& Hidden-state distillation
& 28.1 & 49.7 & 27.7 & 26.7 & 47.4 & 26.6 \\
& \textbf{DiD (Ours)}
& \textbf{31.1} & \textbf{54.5} & \textbf{31.0}
& \textbf{31.1} & \textbf{54.4} & 30.7 \\
\midrule
\multirow{7}{*}{Faster R-CNN}
& \cellcolor[HTML]{d8d8d8}Softmax reference & \cellcolor[HTML]{d8d8d8}33.1 & \cellcolor[HTML]{d8d8d8}56.0 & \cellcolor[HTML]{d8d8d8}34.1 & \cellcolor[HTML]{d8d8d8}33.1 & \cellcolor[HTML]{d8d8d8}56.0 & \cellcolor[HTML]{d8d8d8}34.1 \\
& Full training & 28.9 & 50.0 & 29.1 & 28.6 & 50.0 & 28.6 \\
\cmidrule{2-8}
& Direct copy & 0.3 & 0.7 & 0.2 & 0.2 & 0.4 & 0.1 \\
& Generic distillation & 14.9 & 30.8 & 12.9 & 8.9 & 19.3 & 7.1 \\
& Softmax-generated pseudo-label distillation
& 25.7 & 46.2 & 25.3 & 25.4 & 46.1 & 24.9 \\
& Hidden-state distillation
& 27.1 & 48.1 & 26.9 & 27.2 & 48.1 & 26.6 \\
& \textbf{DiD (Ours)}
& \textbf{29.3} & \textbf{51.1} & \textbf{29.8}
& \textbf{28.9} & \textbf{50.2} & \textbf{29.4} \\
\bottomrule
\end{tabular}
\caption{
\textbf{Label-free Softmax-to-linear conversion on DOTA-v1.5 with ViT-S.}
Full training is a supervised reference; all other adaptation methods use no detection labels.
}
\vspace{-6pt}
\label{tab:main_relu_elu_results}
\end{table*}

\vspace{-3pt}
\subsection{Experimental Setup}

\vspace{-3pt}
\paragraph{Datasets and evaluation.}
We evaluate label-free detector conversion on DOTA-v1.5~\cite{xia2018dota}, a high-resolution aerial-detection benchmark where global token mixing is costly and small objects are common.
Conversion uses the training images only as unlabeled inputs; detection annotations are never used to optimize the student.
The converted detectors are evaluated using mAP, mAP${_{50}}$, and mAP${_{75}}$.
A generalization study on Pascal VOC~\cite{everingham2010pascal}, testing whether the conversion behavior transfers beyond aerial imagery, is provided in the Appendix.

\vspace{-3pt}
\paragraph{Architectures.}
We use RetinaNet~\cite{lin2017focal} and Faster R-CNN~\cite{ren2015faster} with a ViT-S backbone.
The student is initialized by copying all compatible parameters from the trained Softmax detector and replacing only the backbone attention operator.
During conversion, the Softmax teacher, neck/FPN, and detector heads remain frozen, and only the linear-attention student backbone is optimized.
DiD supervises the multi-scale tensors exposed by the frozen neck/FPN to the downstream detector, using the early-heavy interface weighting and plain MSE defaults of Section~\ref{sec:did}.
We evaluate two linear-attention feature maps: $\phi(x)=\mathrm{ReLU}(x)$ following~\cite{han2024bridging}, and $\phi(x)=\mathrm{ELU}(x)+1$ following~\cite{katharopoulos2020transformers}.

\vspace{-3pt}
\paragraph{Conversion protocol.}
Unless otherwise stated, all conversion methods use 40k iterations with AdamW, an initial learning rate of $10^{-3}$, and learning-rate decay at 4k and 32k iterations.
The default augmentation is random horizontal flipping.
Design ablations use a reduced 20k-step schedule.
All experiments are conducted on NVIDIA RTX A6000 GPUs.
Each experiment is repeated three times with different seeds; standard deviations are reported in the Appendix.

\vspace{-3pt}
\subsection{Baselines}
We compare DiD with four label-free alternatives and one supervised reference.
\emph{Direct copy} replaces the attention operator and evaluates the copied parameters without adaptation, following the attention-weight transfer setting of RADLADS~\cite{goldstein2025radlads}.
\emph{Generic distillation} uses LightlyTrain~\cite{lightly-train} to align global teacher-student representations.
\emph{Softmax-generated pseudo-label distillation} uses AutoDistill~\cite{autodistill} to train the student from the frozen teacher's predicted detections.
\emph{Hidden-state distillation} follows RADLADS-style intermediate representation matching across transformer layers.
\emph{Full training} trains the corresponding linear-attention detector using detection labels and is included only as a supervised reference.

\vspace{-3pt}
\subsection{Main Results}

Table~\ref{tab:main_relu_elu_results} establishes three consistent trends:
Firstly, direct copy produces near-zero mAP for both attention kernels and detector families, confirming that the attention operator cannot be replaced without adaptation.
Secondly, generic representation matching and pseudo-label supervision recover only part of the lost detector performance.
Hidden-state distillation is substantially stronger, showing that intermediate representations matter, but it remains below DiD in every setting.
Thirdly, DiD is the strongest label-free method.
For RetinaNet, it reaches 31.1 mAP with both ReLU and ELU, compared with 28.1 and 26.7 for hidden-state distillation.
Its performance is also comparable to the supervised linear-attention references of 30.9 and 30.8 mAP.
We interpret the small numerical differences around the supervised reference as comparable operating points rather than evidence that label-free conversion is superior to supervised training.

Faster R-CNN exhibits the same overall ranking.
DiD reaches 29.3 mAP with ReLU and 28.9 with ELU, outperforming all tested label-free baselines and reaching performance comparable to the corresponding supervised linear-attention references.
The consistency between RetinaNet, an anchor-based one-stage detector, and Faster R-CNN, a proposal-based two-stage detector, indicates that DiD is not tied to a particular downstream detection architecture.

\begin{figure*}[t]
\centering
\begin{minipage}[t]{0.64\linewidth}
  \centering
  \includegraphics[width=\linewidth]
  {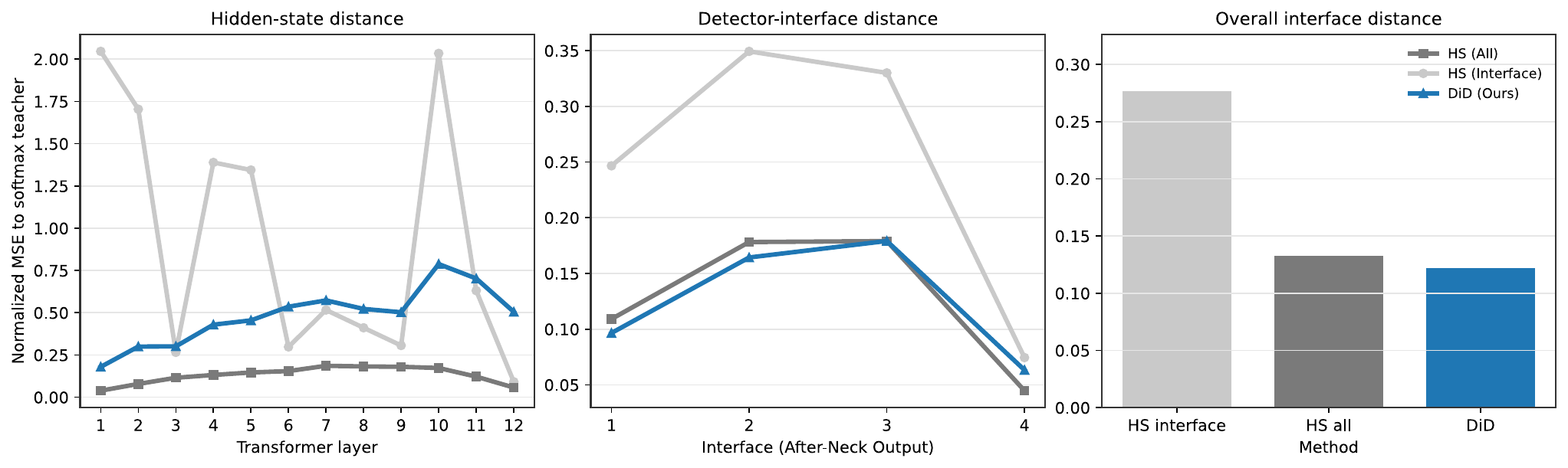}\\
  \vspace{-3pt}
  {\small (a)}
\end{minipage}
\hfill
\begin{minipage}[t]{0.3\linewidth}
  \centering
  \includegraphics[width=\linewidth]
  {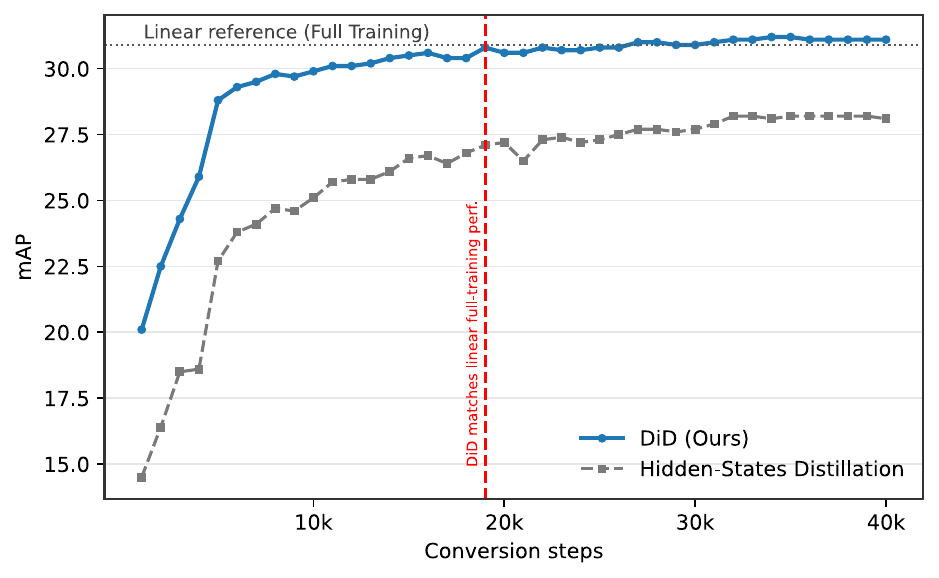}\\
  \vspace{-3pt}
  {\small (b)}
\end{minipage}
\vspace{-6pt}
\caption{
\textbf{Conversion behavior on RetinaNet.}
(a) Hidden-state distillation more closely matches internal transformer states, while DiD more closely matches the detector-facing interface.
(b) DiD consistently outperforms hidden-state distillation during conversion and reaches the supervised linear-training regime after approximately 20k-30k iterations.
}
\vspace{-9pt}
\label{fig:retinanet_combined}
\end{figure*}

\subsection{Interface Alignment and Conversion Dynamics}

To understand why DiD outperforms hidden-state distillation, we compare how the two objectives affect teacher-student alignment inside the backbone and at the detector-facing interface.
Figure~\ref{fig:retinanet_combined}\textcolor{red}{(a)} compares the normalized feature error $\mathrm{nMSE}(s,t)=\mathbb{E}\lVert s-t\rVert_2^2/\mathbb{E}\lVert t\rVert_2^2$ at internal backbone states and detector-facing tensors.

All-layer hidden-state distillation achieves lower mismatch within the transformer, whereas DiD achieves lower mismatch at the detector-facing interface and higher detector accuracy.
This result provides representational evidence for the internal-compensation interpretation: after the attention operator changes, close agreement at every intermediate state is not necessary, provided that the student resolves the discrepancy before its features are consumed by the frozen detector.

Figure~\ref{fig:retinanet_combined}\textcolor{red}{(b)} shows that DiD maintains an advantage over hidden-state distillation throughout conversion and reaches the supervised linear-training regime after approximately 20k-30k steps.
We retain 40k steps as default, while the trajectory shows that most practical recovery occurs earlier.

\subsection{Inference and Conversion Efficiency}

\begin{wraptable}[9]{r}{0.5\textwidth}
\centering
\small
\vspace{-12pt}
\begin{tabular}{lcccc}
\toprule
GPUs & Global batch & Iter/s & \multicolumn{2}{c}{Wall time (min)} \\
\cmidrule(lr){4-5}
& & & 20k & 40k \\
\midrule
1 & 8 & 2.98 & 112 & 224 \\
2 & 8 & 5.20 & 64 & 128 \\
4 & 8 & 7.67 & 44 & 87 \\
\bottomrule
\end{tabular}
\caption{
\textbf{RetinaNet ReLU conversion throughput at fixed global batch size 8.}
}
\vspace{-6pt}
\label{tab:conversion_walltime}
\end{wraptable}

\begin{figure}[t]
\centering
\includegraphics[width=0.7\linewidth]
{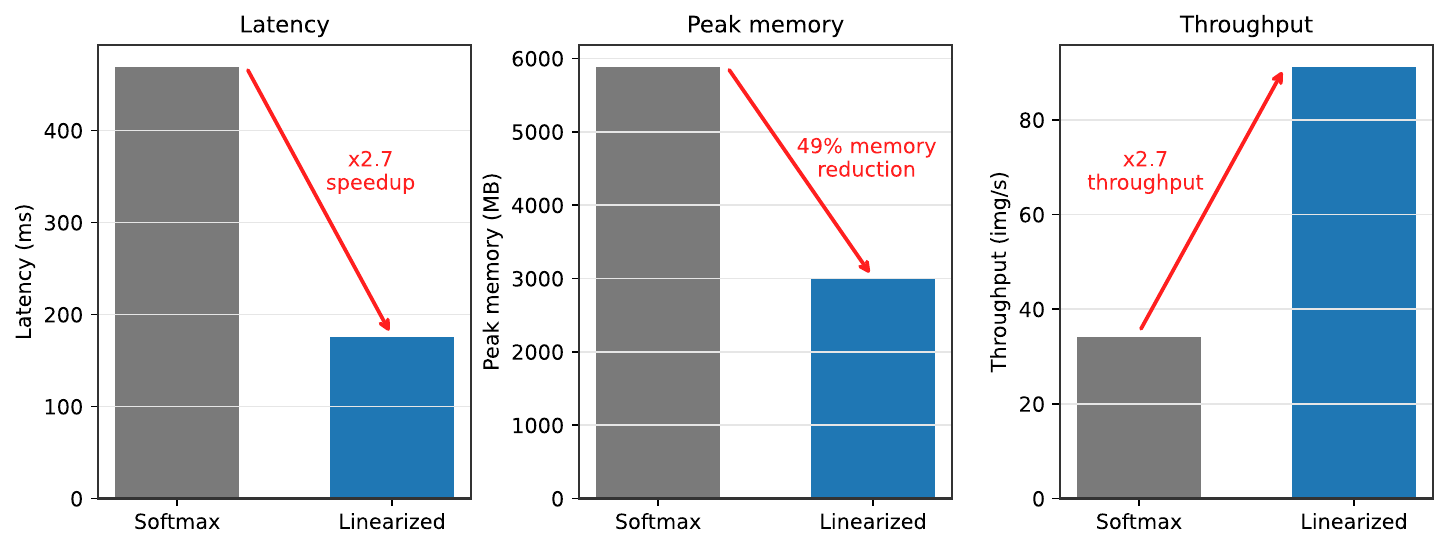}
\vspace{-6pt}
\caption{
\textbf{Backbone-only inference comparison between the Softmax and ReLU-linearized ViT.}
Linear attention substantially reduces latency and peak memory.
}
\vspace{-6pt}
\label{fig:vit_attention_benchmark}
\end{figure}

The converted model realizes the systems benefit that motivates linear attention.
Figure~\ref{fig:vit_attention_benchmark} compares backbone-only inference for Softmax attention and the ReLU-linearized ViT.
The linearized backbone reduces average latency by approximately 62\% and peak memory by approximately 49\%, showing that DiD produces a deployable efficiency improvement rather than only a representational substitution.

The adaptation stage is also short relative to full detector training.
Table~\ref{tab:conversion_walltime} reports conversion throughput at a fixed global batch size of 8: increasing the number of GPUs from one to four raises throughput from 2.98 to 7.67 iterations per second.
The 20k-step practical regime takes approximately 44 minutes on four GPUs, while the conservative 40k-step default takes approximately 87 minutes.

\subsection{Ablation Studies}
\label{sec:ablations}

All ablations use RetinaNet with ReLU linear attention on DOTA-v1.5 and a reduced 20k-step conversion schedule unless otherwise stated.

\begin{wraptable}[7]{r}{0.5\textwidth}
\centering
\small
\vspace{-10pt}
\begin{tabular}{lccc}
\toprule
Interface weighting & mAP & mAP${_{50}}$ & mAP${_{75}}$ \\
\midrule
Uniform & 29.8 & 52.1 & 29.4 \\
Late-heavy & 28.4 & 49.8 & 28.2 \\
Early-heavy & \textbf{30.7} & \textbf{54.1} & \textbf{30.1} \\
\bottomrule
\end{tabular}
\caption{
\textbf{Effect of detector-interface weighting.}
}
\vspace{-3pt}
\label{tab:ablation_weight}
\end{wraptable}

\paragraph{Interface weighting.}
Table~\ref{tab:ablation_weight} compares uniform, late-heavy, and early-heavy weighting over the five detector-facing tensors.
Early-heavy weighting performs best among the evaluated schemes, reaching 30.7 mAP compared with 29.8 for uniform weighting and 28.4 for late-heavy weighting.
This result shows that the five interface levels do not contribute equally under the tested conversion setting.
We therefore adopt early-heavy weighting as the default without claiming that this ordering is universally optimal across detector architectures.

\begin{wraptable}[9]{r}{0.5\textwidth}
\centering
\small
\vspace{-12pt}
\begin{tabular}{lccc}
\toprule
Loss & mAP & mAP${_{50}}$ & mAP${_{75}}$ \\
\midrule
MSE & \textbf{30.7} & \textbf{54.1} & \textbf{30.1} \\
L1 & 30.0 & 52.5 & 29.3 \\
Smooth L1 & 30.3 & 53.0 & 29.7 \\
\midrule
MSE + cosine & 30.4 & 53.2 & 29.8 \\
L1 + cosine & 29.3 & 51.5 & 28.8 \\
Smooth L1 + cosine & 30.4 & 53.1 & 29.9 \\
\bottomrule
\end{tabular}
\caption{
\textbf{Effect of the reconstruction loss.}
}
\vspace{-3pt}
\label{tab:ablation_loss}
\end{wraptable}

\paragraph{Reconstruction loss.}
Table~\ref{tab:ablation_loss} compares MSE, L1, Smooth L1, and their cosine-augmented variants.
Plain MSE performs best, while Smooth L1 remains close and cosine regularization provides no consistent benefit.
This result confirms that DiD does not rely on a specialized reconstruction objective: a simple MSE loss is sufficient once the detector-facing interface has been selected as the conversion target.

\begin{wraptable}[7]{r}{0.64\textwidth}
\centering
\small
\vspace{-10pt}
\begin{tabular}{lccc}
\toprule
Train augmentation & mAP & mAP${_{50}}$ & mAP${_{75}}$ \\
\midrule
RandomHorizontalFlip & \textbf{30.7} & \textbf{54.1} & \textbf{30.1} \\
+ Medium ColorJitter + RandomAffine & 30.0 & 52.2 & 29.8 \\
+ Strong ColorJitter + RandomAffine & 27.0 & 47.9 & 26.4 \\
\bottomrule
\end{tabular}
\caption{
\textbf{Effect of augmentation severity.}
}
\vspace{-3pt}
\label{tab:ablation_augmentation}
\end{wraptable}

\paragraph{Augmentation severity.}
Light detection-style augmentation performs best in Table~\ref{tab:ablation_augmentation}.
Strong color and geometric perturbations reduce mAP, consistent with the purpose of conversion: DiD aligns the student to an already trained detector interface rather than learning invariances from scratch.

\begin{wraptable}[7]{r}{0.5\textwidth}
\centering
\small
\vspace{-10pt}
\begin{tabular}{lccc}
\toprule
Conversion source & mAP & mAP${_{50}}$ & mAP${_{75}}$ \\
\midrule
DOTA-v1.5 & \textbf{30.7} & \textbf{54.1} & \textbf{30.1} \\
FAIR1M & 25.1 & 43.5 & 25.0 \\
ImageNet & 20.2 & 36.6 & 19.7 \\
\bottomrule
\end{tabular}
\caption{
\textbf{Effect of the conversion-data domain.}
}
\vspace{-3pt}
\label{tab:ablation_domain}
\end{wraptable}

\paragraph{Source domain.}
Table~\ref{tab:ablation_domain} compares unlabeled conversion images from DOTA-v1.5, FAIR1M~\cite{sun2022fair1m}, and ImageNet~\cite{russakovsky2015imagenet}.
In-domain DOTA images perform best, followed by the related aerial domain and then generic natural images.
Because supervision comes from the teacher, conversion data is most useful when it exposes the student to the visual distribution that the deployed detector will serve.

\begin{wrapfigure}[11]{r}{0.48\textwidth}
\centering
\vspace{-2pt}
\includegraphics[width=0.9\linewidth]
{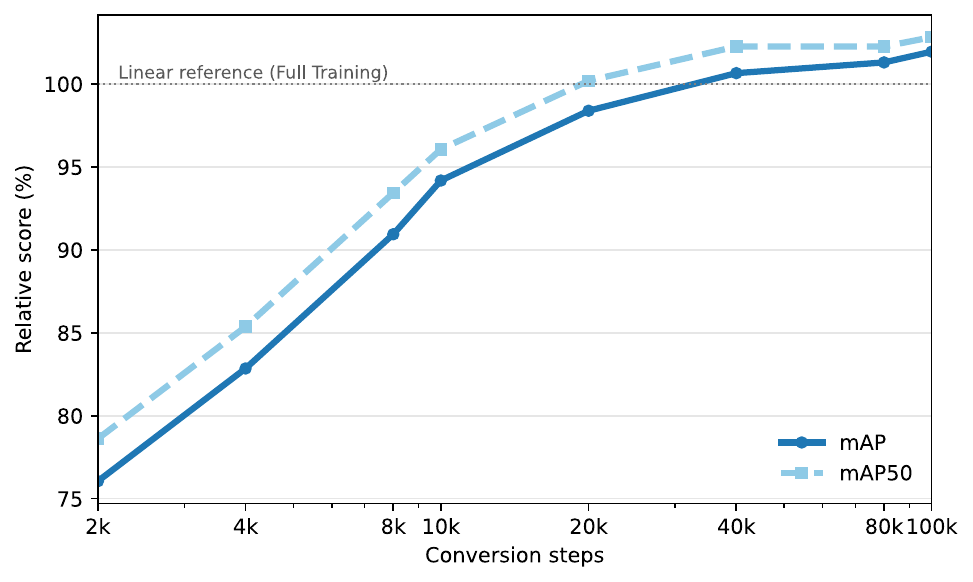}
\vspace{-8pt}
\caption{
\textbf{Scaling behavior of RetinaNet DiD on DOTA-v1.5.}
Scores are normalized to the linear full-training.
}
\vspace{-16pt}
\label{fig:retinanet_scaling}
\end{wrapfigure}

\paragraph{Conversion budget.}
While Figure~\ref{fig:retinanet_combined}\textcolor{red}{(b)} compares the optimization trajectories of DiD and hidden-state distillation, Figure~\ref{fig:retinanet_scaling} focuses specifically on the effect of the DiD conversion budget.
Most of the final performance is recovered within 20k-40k iterations, after which the improvement becomes modest.
This supports the 40k schedule as a conservative default while showing that a shorter 20k conversion already reaches the practical operating regime.

\vspace{1.5pt}
More ablations are provided in the Appendix.

\section{Conclusion}
We presented Detector-Interface Distillation (DiD), a label-free method for converting trained Softmax-attention ViT detectors into efficient linear-attention ones.
Our central finding is that successful conversion is governed by compatibility at the detector-facing interface: supervising the tensors actually consumed by the frozen detector lets the linear backbone compensate internally, where direct operator replacement and generic distillation fail.
On DOTA-v1.5, DiD is the strongest label-free conversion method for both RetinaNet and Faster R-CNN, reaches an operating point comparable to fully supervised linear-attention training, and completes in about 87 minutes on 4 GPUs while substantially reducing inference latency and peak memory.
We hope DiD offers a simple post-training route to efficient ViT detectors, and that interface-aware objectives extend beyond the ViT-S scale studied here to larger backbones, additional detector families, and broader architecture-conversion settings.

{
\small
\bibliographystyle{unsrt}
\bibliography{sections/11_references}
}

\newpage
\appendix \appendix

\cftsetindents{section}{0em}{2.1em}
\cftsetindents{subsection}{2.1em}{2.9em}
\cftsetindents{subsubsection}{5.0em}{3.8em}
\setlength{\cftbeforesecskip}{6pt}
\setlength{\cftbeforesubsecskip}{2pt}
\setlength{\cftbeforesubsubsecskip}{2pt}

\setcounter{section}{0}
\setcounter{table}{0}
\setcounter{figure}{0}
\setcounter{equation}{0}

\ifreview \nolinenumbers \fi
\begin{center}
\Large \textbf{Technical Appendices and Supplementary Material} \\
\end{center}
\ifreview \linenumbers \fi


\renewcommand{\thesection}{\Alph{section}}
\renewcommand{\thetable}{\Alph{table}}
\renewcommand{\thefigure}{\Alph{figure}}
\renewcommand{\theequation}{\Alph{equation}}

\section{Baseline Objectives}
\label{sec:appendix_baselines}

This section formalizes the optimization-based baseline objectives used in the main paper.
Let $\theta_t$ and $\theta_s$ denote the teacher and student parameters, respectively.

\paragraph{Generic distillation.}
Our first optimization baseline adopts a generic teacher-student distillation objective on unlabeled images, instantiated with the LightlyTrain framework.
The teacher is kept fixed, and the student is optimized as
\begin{equation}
\theta_s^\star
=
\arg\min_{\theta_s}
\mathbb{E}_{x \sim \mathcal{D}}
\left[
\mathcal{L}_{\mathrm{ssl}}(x; \theta_s, \theta_t)
\right].
\end{equation}
Concretely, the loss matches global representations across two stochastic augmentations $x^{(1)}$ and $x^{(2)}$ of the same unlabeled image:
\begin{equation}
\mathcal{L}_{\mathrm{ssl}}
=
\mathcal{C}_{\mathrm{vec}}\left(
g_s(x^{(1)}),
g_t(x^{(2)})
\right)
+
\mathcal{C}_{\mathrm{vec}}\left(
g_s(x^{(2)}),
g_t(x^{(1)})
\right),
\end{equation}
where $g_t(\cdot)$ and $g_s(\cdot)$ denote the teacher and student global projection heads, and
\begin{equation}
\mathcal{C}_{\mathrm{vec}}(a,b)
=
1 - \frac{\langle a, b \rangle}{\lVert a\rVert_2 \, \lVert b\rVert_2}.
\end{equation}
This baseline preserves high-level semantic consistency, but it does not explicitly constrain the detector-facing interface tensors consumed by the downstream detector.
In practice, we use the LightlyTrain~\cite{lightly-train} implementation and follow its recommended setting.

\paragraph{Softmax-generated pseudo-label distillation.}
The next baseline uses the frozen Softmax detector as a pseudo-label teacher.
For each unlabeled image $x$, the teacher produces pseudo detections $\hat{y}_t(x)$, and the student is trained with the corresponding detection loss:
\begin{equation}
\theta_s^\star
=
\arg\min_{\theta_s}
\mathbb{E}_{x \sim \mathcal{D}}
\left[
\mathcal{L}_{\mathrm{det}}\left(f_s(x), \hat{y}_t(x)\right)
\right].
\end{equation}
This baseline transfers detector supervision at the output level rather than through intermediate feature matching.
It is more task-specific than generic self-supervised distillation, but it still does not directly enforce compatibility at the deployed detector-facing interface tensors.
We implement this baseline with the AutoDistill~\cite{autodistill} framework.

\paragraph{Hidden-state distillation.}
A more direct representation-matching baseline supervises backbone hidden states across transformer layers, following the general strategy of RADLADS~\cite{goldstein2025radlads}.
Let $\mathcal{K}$ denote the set of supervised hidden-state layers, and let $h_t^{(\ell)}(x)$ and $h_s^{(\ell)}(x)$ denote the teacher and student hidden states at layer $\ell$.
Let $\alpha_\ell > 0$ be the weight assigned to layer $\ell \in \mathcal{K}$.
In our implementation, $\alpha_\ell$ may differ between detector-interface layers and non-interface layers, although we use equal weighting by default.
The loss is
\begin{equation}
\begin{aligned}
\mathcal{L}_{\mathrm{hid}}
&=
\frac{1}{\sum_{\ell \in \mathcal{K}} \alpha_\ell}
\sum_{\ell \in \mathcal{K}}
\alpha_\ell
\Bigl(
\lVert h_s^{(\ell)}(x) - h_t^{(\ell)}(x) \rVert_F^2+
\lambda_{\mathrm{cos}}
\mathcal{C}_{\mathrm{tok}}\left(
h_s^{(\ell)}(x),
h_t^{(\ell)}(x)
\right)
\Bigr),
\end{aligned}
\end{equation}
where $\lVert\cdot\rVert_F$ is the Frobenius norm and
\begin{equation}
\mathcal{C}_{\mathrm{tok}}(a,b)
=
\frac{1}{|\Omega(a)|}
\sum_{u \in \Omega(a)}
\left(
1
-
\frac{\langle a_u, b_u \rangle}
{\lVert a_u\rVert_2 \, \lVert b_u\rVert_2}
\right)
\end{equation}
is a tokenwise cosine alignment loss, with $\Omega(a)$ denoting the set of token locations in the hidden-state tensor.
This objective encourages the converted student to preserve the teacher's intermediate transformer behavior in a generic, detector-agnostic way.

These baselines test whether global semantic distillation, output-level pseudo-label supervision, or hidden-state imitation suffice for Softmax-to-linear conversion in object detection.

\section{Variance Across Seeds}
\label{sec:appendix_variance}

Each experiment is repeated three times with different random seeds.
Table~\ref{tab:appendix_variance} reports the mean and standard deviation over the three seeds for DiD with RetinaNet and Faster R-CNN under the ReLU and ELU kernels on DOTA-v1.5.

\begin{table}[t]
\centering
\small
\setlength{\tabcolsep}{3pt}
\begin{tabular}{lcccc}
\toprule
Detector & Kernel & mAP & mAP${_{50}}$ & mAP${_{75}}$ \\
\midrule
\multirow{2}{*}{RetinaNet}
& ReLU & 31.1 $\pm$ 0.1 & 54.5 $\pm$ 0.2 & 31.0 $\pm$ 0.1 \\
& ELU  & 31.1 $\pm$ 0.1 & 54.4 $\pm$ 0.1 & 30.7 $\pm$ 0.1 \\
\midrule
\multirow{2}{*}{Faster R-CNN}
& ReLU & 29.3 $\pm$ 0.2 & 51.1 $\pm$ 0.2 & 29.8 $\pm$ 0.2 \\
& ELU  & 28.9 $\pm$ 0.1 & 50.2 $\pm$ 0.1 & 29.4 $\pm$ 0.2 \\
\bottomrule
\end{tabular}
\vspace{3pt}
\caption{
\textbf{Mean $\pm$ standard deviation over three seeds for DiD conversion on DOTA-v1.5 with a ViT-S backbone.}
}
\vspace{-6pt}
\label{tab:appendix_variance}
\end{table}

\section{Ablation on Teacher Initialization}
\label{sec:appendix_teacher_init}

We test ImageNet initialization, DINO initialization~\cite{caron2021emerging}, and detector training without backbone pretraining.
Table~\ref{tab:ablation_pretraining} shows that DiD remains stronger than hidden-state distillation for all three teachers.
The final converted accuracy nevertheless depends on the interface learned by the original detector: stronger Softmax teachers generally yield stronger converted models.

\begin{table*}[t]
\centering
\small
\setlength{\tabcolsep}{3pt}
\begin{tabular}{lccc|ccc|ccc}
\toprule
\multirow{2}{*}{Method}
& \multicolumn{3}{c}{ImageNet}
& \multicolumn{3}{c}{DINO}
& \multicolumn{3}{c}{No pretraining} \\
& mAP & mAP${_{50}}$ & mAP${_{75}}$
& mAP & mAP${_{50}}$ & mAP${_{75}}$
& mAP & mAP${_{50}}$ & mAP${_{75}}$ \\
\midrule
Softmax reference
& 33.5 & 58.1 & 33.2
& 31.4 & 53.8 & 31.5
& 25.9 & 46.5 & 25.2 \\
\midrule
Hidden-state distillation
& 28.1 & 49.7 & 27.7
& 28.7 & 49.7 & 28.5
& 26.0 & 46.4 & 25.3 \\
DiD
& \textbf{31.1} & \textbf{54.5} & \textbf{31.0}
& \textbf{31.0} & \textbf{53.1} & \textbf{31.5}
& \textbf{26.2} & \textbf{46.7} & \textbf{25.4} \\
\bottomrule
\end{tabular}
\caption{
\textbf{Compatibility with different teacher initializations for RetinaNet ReLU conversion on DOTA-v1.5.}
}
\vspace{-6pt}
\label{tab:ablation_pretraining}
\end{table*}

In the no-pretraining setting, DiD slightly exceeds the Softmax reference by 0.3 mAP.
We treat this small difference as a setting-specific result rather than evidence that conversion generally improves the teacher.
Overall, DiD is compatible with all tested teacher initializations, and its converted accuracy inherits the operating point of the learned detector interface.

\section{Generalization to Pascal VOC}
\label{sec:appendix_voc}

Table~\ref{tab:main_relu_elu_results_voc} applies the same conversion protocol and baselines as the main DOTA-v1.5 study to Pascal VOC~\cite{everingham2010pascal} to test whether the conversion behavior transfers beyond aerial imagery.

\begin{table*}[t]
\centering
\small
\setlength{\tabcolsep}{3pt}
\begin{tabular}{llccc|ccc}
\toprule
\multirow{2}{*}{Detector} & \multirow{2}{*}{Method}
& \multicolumn{3}{c}{ReLU}
& \multicolumn{3}{c}{ELU} \\
& & mAP & mAP${_{50}}$ & mAP${_{75}}$ & mAP & mAP${_{50}}$ & mAP${_{75}}$ \\
\midrule
\multirow{7}{*}{RetinaNet}
& \cellcolor[HTML]{d8d8d8}Softmax reference
& \cellcolor[HTML]{d8d8d8}66.6 & \cellcolor[HTML]{d8d8d8}90.3 & \cellcolor[HTML]{d8d8d8}75.2
& \cellcolor[HTML]{d8d8d8}66.6 & \cellcolor[HTML]{d8d8d8}90.3 & \cellcolor[HTML]{d8d8d8}75.2 \\
& Full training
& 60.3 & 85.5 & 67.4
& 59.4 & 84.8 & 66.2 \\
\cmidrule{2-8}
& Direct copy
& 3.2 & 7.0 & 1.8
& 2.1 & 5.0 & 1.0 \\
& Generic distillation
& 46.5 & 75.8 & 49.2
& 38.0 & 67.1 & 37.2 \\
& Softmax-generated pseudo-label distillation
& 54.0 & 82.2 & 58.4
& 55.6 & 83.5 & 60.7 \\
& Hidden-state distillation
& 57.5 & 84.3 & 63.8
& 56.1 & 83.2 & 61.9 \\
& \textbf{DiD (Ours)}
& \textbf{60.8} & \textbf{86.2} & \textbf{68.0}
& \textbf{59.7} & \textbf{85.2} & \textbf{66.6} \\
\midrule
\multirow{7}{*}{Faster R-CNN}
& \cellcolor[HTML]{d8d8d8}Softmax reference
& \cellcolor[HTML]{d8d8d8}66.0 & \cellcolor[HTML]{d8d8d8}89.8 & \cellcolor[HTML]{d8d8d8}75.8
& \cellcolor[HTML]{d8d8d8}66.0 & \cellcolor[HTML]{d8d8d8}89.8 & \cellcolor[HTML]{d8d8d8}75.8 \\
& Full training
& 57.6 & 83.4 & 64.5
& 56.7 & 82.7 & 63.1 \\
\cmidrule{2-8}
& Direct copy
& 0.7 & 1.6 & 0.3
& 0.4 & 1.0 & 0.1 \\
& Generic distillation
& 44.5 & 74.5 & 46.4
& 35.8 & 64.0 & 34.6 \\
& Softmax-generated pseudo-label distillation
& 54.3 & 82.4 & 59.7
& 53.7 & 81.9 & 58.4 \\
& Hidden-state distillation
& 55.4 & 82.6 & 61.3
& 55.0 & 82.1 & 60.7 \\
& \textbf{DiD (Ours)}
& \textbf{58.0} & \textbf{84.1} & \textbf{65.3}
& \textbf{56.9} & \textbf{83.2} & \textbf{63.8} \\
\bottomrule
\end{tabular}
\caption{
\textbf{Label-free Softmax-to-linear conversion on Pascal VOC with ViT-S.}
Full training is a supervised reference; all other adaptation methods use no detection labels.
}
\vspace{-6pt}
\label{tab:main_relu_elu_results_voc}
\end{table*}

\end{document}